\documentclass[runningheads]{llncs}
\usepackage{graphicx}

\usepackage{tikz}
\usepackage{comment}
\usepackage{amsmath,amssymb} 
\usepackage{color}
\usepackage{multirow}
\usepackage[ruled,vlined]{algorithm2e}
\usepackage[pagebackref,breaklinks,colorlinks]{hyperref}
\usepackage[capitalize]{cleveref}
\Crefname{section}{Sec.}{Secs.}
\Crefname{table}{Tab.}{Tabs.}

\usepackage[accsupp]{axessibility}  

\begin{document}
\pagestyle{headings}
\mainmatter
\def\ECCVSubNumber{6982}  

\title{Weakly-Supervised Temporal Action Detection for Fine-Grained Videos with Hierarchical Atomic Actions} 

\titlerunning{Weakly-Supervised Temporal Action Detection for Fine-Grained Videos}
%
\author{Zhi Li\inst{1} \and
Lu He\inst{2} \and
Huijuan Xu\inst{3}}
\authorrunning{Z. Li, L. He, H. Xu}
%
\institute{University of California, Berkeley, USA \\ 
\email{zhili@berkeley.edu}
\and
Tencent America, Palo Alto, USA\\ 
\email{lhluhe@tencent.com}
\and
Pennsylvania State University, University Park, USA\\
\email{hkx5063@psu.edu}}

\maketitle

\begin{abstract}

Action understanding has evolved into the era of fine granularity, as most human behaviors in real life have only minor differences.
To detect these fine-grained actions accurately in a label-efficient way, we tackle the problem of weakly-supervised fine-grained temporal action detection in videos for the first time.
Without the careful design to capture subtle differences between fine-grained actions, previous weakly-supervised models for general action detection cannot perform well in the fine-grained setting.
We propose to model actions as the combinations of reusable atomic actions which are automatically discovered from data through self-supervised clustering, in order to capture the commonality and individuality of fine-grained actions.
The learnt atomic actions, represented by visual concepts, are further mapped to fine and coarse action labels leveraging the semantic label hierarchy. Our approach constructs a visual representation hierarchy of four levels: clip level, atomic action level, fine action class level and coarse action class level, with supervision at each level.
Extensive experiments on two large-scale fine-grained video datasets, FineAction and FineGym, show the benefit of our proposed weakly-supervised model for fine-grained action detection, and it achieves state-of-the-art results.


\end{abstract}

\section{Introduction}
\label{sec:intro}

\begin{figure}[t]
  \centering
   \includegraphics[width=0.70\linewidth]{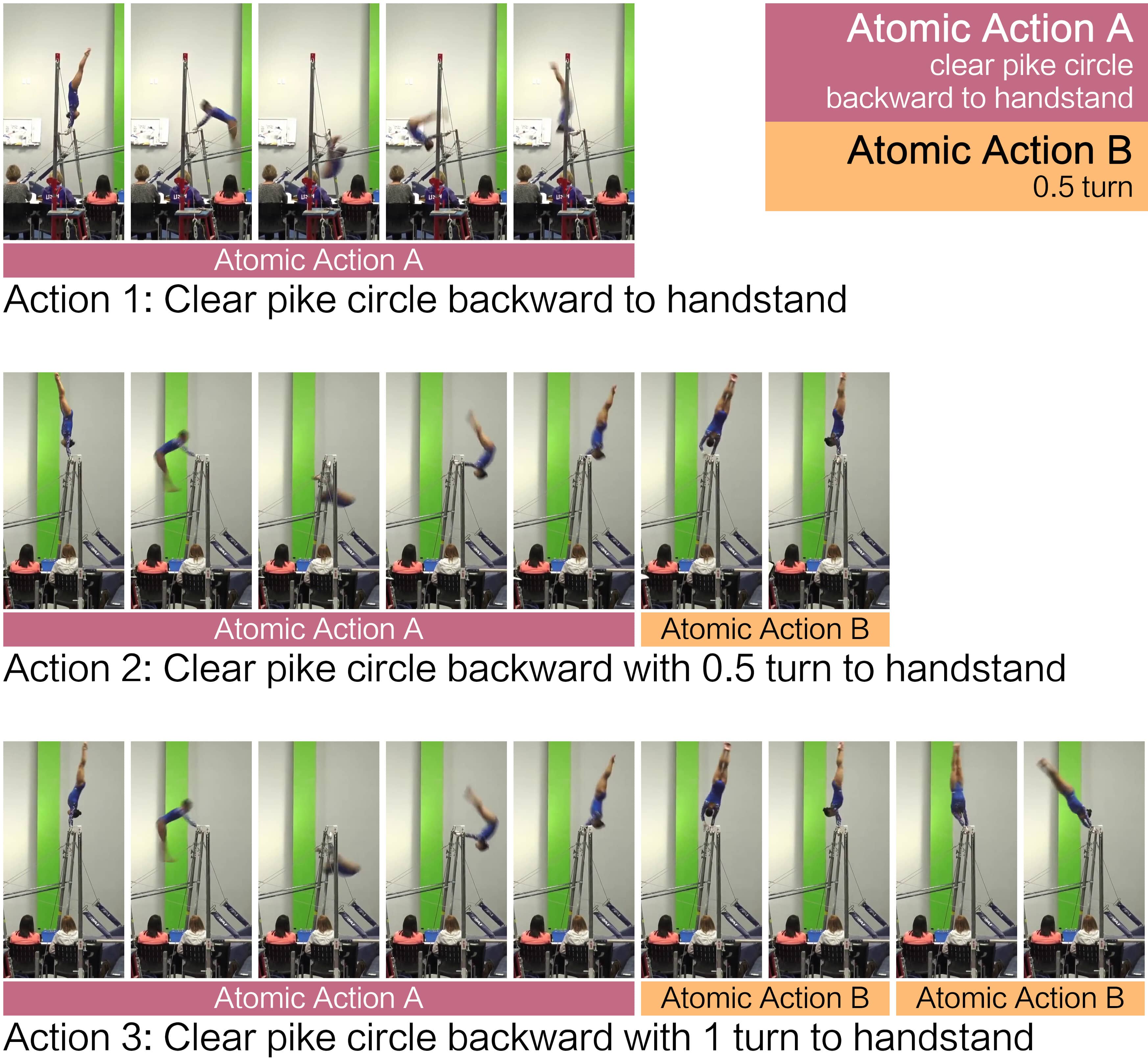}
   \caption{Fine-grained action examples from FineGym dataset \cite{shao2020finegym} with atomic action decomposition. The three actions all share the same atomic action A (\textit{clear pike circle backward to handstand}), but differ in the ending phase of the action. Action 1 only contains atomic action A. Action 2 has another atomic action B (\textit{0.5 turn}) following A, and action 3 has atomic action B repeated twice after A}
   \label{fig:atomic}
\end{figure}

Early video benchmarks \cite{jiang2014thumos,caba2015activitynet} mostly contain actions with distinct movement. Recent works have evolved into fine-grained action understanding \cite{shao2020finegym,liu2021fineaction}, which is closer to the distribution of actions in real life. These fine-grained actions are either visually similar actions, e.g., \textit{mop floor} and \textit{sweep floor}, or actions from continuously recorded instructional videos. Designing algorithms to detect fine-grained actions could potentially assist people in acquiring new skills, since people usually learn from continuous instructional videos, and correctly segmented action steps could benefit their learning process. Fine-grained action detection algorithms are also needed on home assistance devices to better perceive actions happening in home and take appropriate actions to assist people.

For temporal action detection, data annotation in the form of start and end frames suffers from high annotation cost due to the large volume of videos, and annotation consistency issue among various annotators. These problems are even more severe in fine-grained videos, because the distinction between fine-grained actions is not so obvious, and the time granularity of action annotation is also more refined. To alleviate these problems and conduct label-efficient temporal action detection in fine-grained videos, we propose to tackle weakly-supervised action detection with only video-level action labels and without temporal annotations of when the actions take place, for the first time in fine-grained videos.

Traditional weakly-supervised action detection models are mostly based on Multiple Instance Learning (MIL) \cite{MIL_summary}. The core of these MIL-based models is video-level action classification, and video segments having high classification activation are selected as detected action segments. Fine-grained videos, however, introduce extra challenges in MIL, as fine-grained actions are visually similar and the differences only manifest in small details (examples are shown in \cref{fig:atomic}). As a result, classification itself is much more difficult in fine-grained tasks~\cite{liu2021fineaction,shao2020finegym}.

Prior works \cite{ji2020action,gaidon2013temporal,lillo2014discriminative} on action detection decompose human's behaviours into a series of atomic building blocks inspired by human's cognition of actions. Following this idea, we model fine-grained action details through the lens of \textbf{Atomic Actions}, defined as short temporal parts representing a single semantically meaningful component of an action, to benefit their classification. This is inspired by the observation that visually similar fine-grained actions often share common atomic actions and only differ in the key atomic segments. For example, as shown in \cref{fig:atomic}, when comparing the three fine-grained actions, \textit{clear pike circle backward to handstand}, \textit{clear pike circle backward with 0.5 turn to handstand}, and \textit{clear pike circle backward with 1 turn to handstand}, we humans perceive their differences by looking at the last atomic action. However, the boundaries of atomic actions are not explicitly labeled, and are very hard to obtain even if we want to label them.

Inspired by \textbf{Visual Concept} in fine-grained image understanding \cite{ghorbani2019automating,higgins2017scan,chen2019looks}, we propose to automatically discover these atomic actions in the feature level, which are hard to be defined in advance by humans. This is different from prior works defining atomic building blocks ahead of time \cite{ji2020action,gaidon2013temporal,lillo2014discriminative}. 
We leverage self-supervised clustering to discover temporal visual concepts from video clip features. We then use visual concepts to represent atomic actions and model a fine-grained action as a composition of atomic actions. Specifically, MIL classifier weights are used to select visual concepts for each fine-grained action. Visual concepts capture fine details, and can in turn facilitate the MIL training.

The MIL classification relies on the most discriminative part of the action and thus is prone to detecting only the discriminative segment as target action. 
To learn the commonality among fine-grained actions and detect complete action, we incorporate coarse-to-fine label hierarchy often available in fine-grained videos to regularize the commonality between actions in the feature space.
Concretely, we aggregate fine-level visual concepts to obtain coarse-level visual concept representations, and connect them with the coarse-level labels' supervision.

In summary, we propose a Hierarchical Atomic Action Network (HAAN) to model the commonality and individuality of fine-grained actions in the MIL framework to conduct weakly-supervised fine-grained temporal action detection\footnote{Code is available at \url{https://github.com/lizhi1104/HAAN.git}}. The contributions of this paper include:
\begin{enumerate}
    \item We are the first to tackle the weakly-supervised fine-grained temporal action detection problem, and benchmark previous weakly-supervised approaches for general action detection on fine-grained datasets.
    \item Propose a self-supervised learning approach to discover visual concepts for building atomic actions, which promotes the learning of fine action details.
    \item Leverage coarse-to-fine semantic label relationship to construct a hierarchical visual concept system to encourage learning commonality among fine actions.
    \item Conduct extensive experiments for the HAAN model on two large-scale fine-grained video datasets: FineAction and FineGym, and achieve state-of-the-art results.  

\end{enumerate}

\section{Related Work}
\subsection{Weakly-Supervised Temporal Action Detection}

Labeling an action's start and end points in video frames is a labor-intensive task. Thus the weakly-supervised approach, which drastically reduces annotation requirements, becomes an important research topic. Under weakly-supervised setting, detection models are trained with only video-level action labels, without temporal annotations of the action's start and end points. Previous research mostly involved the Multiple Instance Learning (MIL) \cite{carbonneau2018multiple,MIL_summary,DIETTERICH199731} process. MIL leverages classification signals to help action detection: given a video-level action label, some parts of the video must contain the specific action, and the key is to figure out which parts. To solve that, MIL conducts classification for the whole video, during which it carefully selects some parts of the video with high activation scores as action segments. Typically the selection method can be either k-max pooling \cite{wang2017untrimmednets,paul2018w,narayan20193c,shou2018autoloc,ma2021weakly}, or attention-based pooling \cite{wang2017untrimmednets,nguyen2018weakly,yuan2019marginalized,liu2019completeness}.

On top of the basic MIL classifier, previous researchers also explored adding different regularization losses, constraints or temporal property modeling to boost the action detection performance. Nguyen et al. \cite{nguyen2018weakly} added a sparsity loss to the attention weights in order to predict accurate action segments. Paul et al. \cite{paul2018w} proposed a co-activity similarity loss to enforce similarity between features of the same action. Shou et al. \cite{shou2018autoloc} added a contrastive loss to help action boundary prediction. And Ma et al. \cite{ma2021weakly} introduced a class-agnostic actionness score, which leverages context to help focus on the parts that contain actions.

None of these prior works, however, tackled fine-grained actions. Two widely used benchmarks, THUMOS-14 \cite{jiang2014thumos} and ActivityNet-1.2 \cite{caba2015activitynet}, do not consistently contain fine-grained actions, either. Sun et al. \cite{sun2015temporal} mentioned fine-grained actions in weakly-supervised setting, but their definition of fine-grained actions is closer to the granularity of THUMOS-14 and ActivityNet-1.2, and they require extra supervision from web images. To the best of our knowledge, we are the first to tackle weakly-supervised action detection task for fine-grained actions. 

\subsection{Fine-Grained Temporal Action Detection}
Fine-grained action detection has become an increasingly important research topic recently. Several fine-grained action datasets have been proposed, such as MPII \cite{rohrbach2012database}, Salad 50 \cite{stein2013combining}, MERL Shopping \cite{singh2016multi}, GTEA \cite{fathi2011learning}, EPIC-KITCHENS-100 \cite{damen2022rescaling}, FineGym \cite{shao2020finegym}, and FineAction \cite{liu2021fineaction}. Small differences between fine-grained action categories post extra challenges for action detection. As a result, it is not uncommon to see that models designed for general action detection perform much worse on fine-grained datasets \cite{shao2020finegym,liu2021fineaction}.

Previous works explored different methods to model fine-grained action details. Mac et al. \cite{deform} applied deformable convolution to extract local spatio-temporal features in order to obtain fine-grained motion details. Piergiovanni and Ryoo \cite{FineBaseball} leveraged temporal structures to detect actions. Mavroudi et al.'s work \cite{random_field} involved learning a dictionary for action primitives. These primitives capture fine-grained action details, and inspired our work in the use of atomic actions. The most relevant to our work is \cite{ni2014multiple}, which decomposed the detection task into two steps on coarse and fine levels respectively. This inspired our hierarchical modeling of visual concepts. All these prior works require full supervision of action start/end point annotations.

Some previous works' settings are close to weakly-supervised setting, but still require extra supervision other than just the set of actions, such as the order of actions \cite{richard2017weakly,ghoddoosian2022hierarchical}. This research is set to address weakly-supervised action detection for fine-grained videos with only the set of actions available.

\subsection{Atomic Actions and Visual Concepts}

The concept of atomic actions naturally arises when researchers analyze humans' cognitive perception of actions. Gaidon et al. \cite{gaidon2013temporal} used a histogram of features representing atomic actions to conduct action detection. Similarly, Lillo et al. \cite{lillo2014discriminative} modeled human activities as smaller individual components. Recently, Ji et al. \cite{ji2020action} decomposed an action event into atomic spatial-temporal scene graphs, aiming to understand actions as sequences of atomic interactions with the surrounding environments.

Visual concept, on the other hand, is often related to promoting models' interpretability. Chen et al. \cite{chen2016infogan} showed that InfoGAN encodes disentangled visual concepts such as pose, emotion, and hairstyle. Whitney et al. \cite{continuation_learning} learned symbolic representations from video frames. Kim et al. \cite{kim2018interpretability} analyzed how each visual concept of the input image affects the classification prediction. Ghorbani et al. \cite{ghorbani2019automating} proposed a method that automatically recovers visual concepts and determines their importance without user's input. Chen et al. \cite{chen2019looks} proposed to model prototypical parts in image recognition and combine those parts in final predictions. Those prototypical parts can be viewed as visual concepts.

Visual concepts are typically obtained via unsupervised learning \cite{ghorbani2019automating,higgins2017scan}. Given that atomic actions are not directly annotated, these works inspire us to use visual concepts to model atomic actions. In the image field, visual concepts usually correspond to small patches of the image for a single subject \cite{chen2019looks}. In our work, we extend this concept to temporal visual concepts in videos.

\section{Method}
\begin{figure*}[t]
  \centering
  \includegraphics[width=\linewidth]{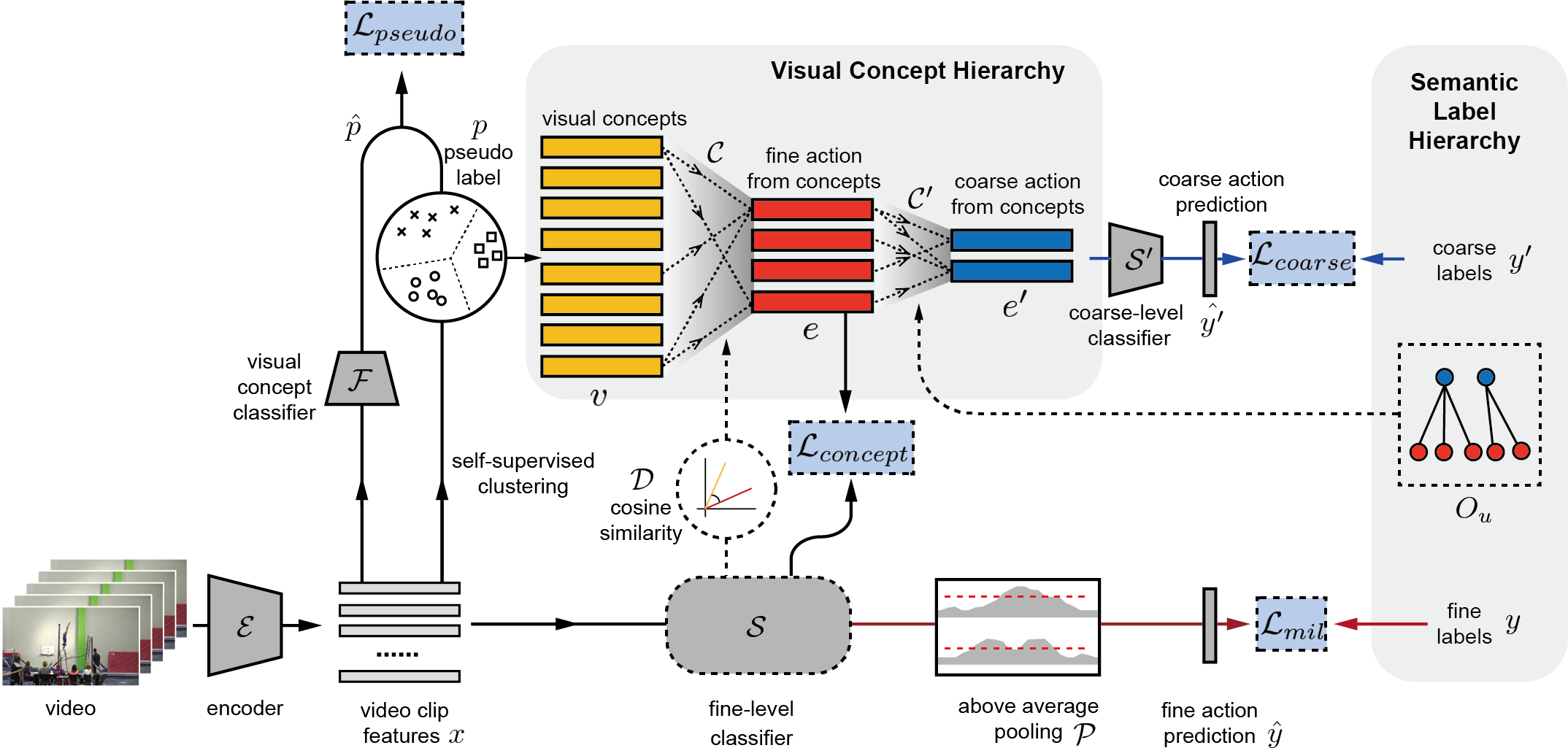}
  \caption{An overview of our HAAN model. It contains a fine-grained action MIL classifier $\mathcal{S}$ trained with $\mathcal{L}_{mil}$. To capture fine-grained details, HAAN leverages self-supervised clustering ($\mathcal{L}_{pseudo}$) to learn visual concepts which represent atomic actions. In order to model each fine-grained action as a composition of atomic actions, we connect visual concept $v$ with the MIL classifier $\mathcal{S}$ based on a distance function $\mathcal{D}$. The model learns to compose visual concepts into fine-grained actions through $\mathcal{L}_{concept}$. In addition, we use the coarse-to-fine semantic label hierarchy (represented by $O_u$) to further compose fine-grained actions into coarse-grained actions, and train it with $\mathcal{L}_{coarse}$}
  \label{fig:model}
\end{figure*}

We first introduce the notations. Given a set of videos with $C$ categories, each video contains $M$ action segments ($M$ can be different for different videos). The $m^{th}$ action segment is described by its class label $a_m\in \{1,\cdots,C\}$ and its start/end timestamp $(Start_m,End_m)$. Under the weakly-supervised action detection setting, for each video we only have access to the set of action labels $Set(a_1,\cdots,a_M)$ at training time. The number of action segments $M$, the order of action segments, and their temporal timestamps are not available. In other words, the annotation for each video can be represented by a multi-hot encoding label vector $y = \{y^1,\cdots,y^C\} \in \{0,1\}^C$, where each entry is a binary indicator representing whether the video contains the corresponding action. The goal is to detect the start and end points $(Start_m,End_m)$ for every action. \cref{fig:model} is an overview of the HAAN model.

\subsection{Multiple Instance Learning}

Multiple Instance Learning (MIL) is widely used in previous non fine-grained weakly-supervised action detection models. The idea is quite straightforward: given the multi-hot video-level action label $y\in \{0,1\}^C $, we know that the video contains specific actions. In order to detect which parts (a.k.a. instances) correspond to that action, MIL uses the signal from action classification loss. The model aggregates information along the temporal axis to obtain a global representation for the whole video, and is trained with classification loss. During this process, the parts that contribute the most to video-level class prediction are selected as target actions.

Following prior work \cite{wang2017untrimmednets,paul2018w,narayan20193c,shou2018autoloc,ma2021weakly,luo2020weakly}, we divide a long video into $T$ short video clips. Then we use feature encoder $\mathcal{E}$ to extract clip level features $\{x_1,\cdots,x_T\}$, where each $x_i\in R^d$ is the $d$-dimensional feature vector for $i^{th}$ clip. A classifier $\mathcal{S}$ then predicts a score for each clip: $s_i=\mathcal{S}(x_i)\in R^C,1\leq i\leq T$. After that, a pooling method $\mathcal{P}$ aggregates scores from all clips $s=\{s_1,s_2,\cdots,s_T\}\in R^{T\times C}$ into a global action label prediction $\hat{y}=\mathcal{P}(s)\in R^C$. $\hat{y}$ is then compared with $y$ via standard classification loss. 

Previous works explored different pooling methods $\mathcal{P}$ including k-max pooling \cite{wang2017untrimmednets,paul2018w,narayan20193c,shou2018autoloc,ma2021weakly} and attention-based pooling \cite{wang2017untrimmednets,nguyen2018weakly,yuan2019marginalized,liu2019completeness}. In our paper, we filter the scores $\{s_1,s_2,\cdots,s_T\}$ with the mean  $\overline{s}$ of all the scores as threshold, and take the average of the scores above the mean. Concretely, for each action class $j$ ($1\leq j\leq C$):

\begin{equation}
  \hat{y^j}=\frac{1}{\sum\limits_{i=1}^T\mathbf{1}(s_i^j\geq \overline{s^j})}\sum_{i=1}^T\mathbf{1}(s_i^j\geq \overline{s^j} )\cdot s_i^j
  \label{eq:mil_pred}
\end{equation}

where $\overline{s^j}$ is the mean of scores of class $j$ for all $T$ clips. The aggregated video-level prediction logit $\hat{y^j}$ is then used in the binary cross entropy ($BCE$) MIL loss with logits. 


\begin{equation}
  \mathcal{L}_{mil} = BCE(\hat{y^j}, y^j)
  \label{eq:l_mil}
\end{equation}

\subsection{Visual Concept Learning}

\label{subsec:visual_concept}
To capture fine-grained temporal details, we introduce self-supervised clustering to discover visual concepts from video features. Before each epoch, we collect features of all clips in the whole training data, and then conduct K-means clustering \cite{macqueen1967classification} on the whole feature pool to obtain $N$ clusters. This generates a pseudo label $p_i$ for each video clip $i$, $p_i \in \{1,\cdots,N\} $ is assigned by the clustering algorithm.

Pseudo label is then used to train a visual concept classifier $\mathcal{F}$. Visual concept classifier $\mathcal{F}$ maps the shared features $x_i$ to a predicted cluster logit, $\hat{p_i}=\mathcal{F}(x_i)\in R^N$. The model then trains $\hat{p_i}$ with cross entropy ($CE$) loss.

\begin{equation}
  \mathcal{L}_{pseudo} = CE(\hat{p_i}, p_i)
  \label{eq:l_pseudo}
\end{equation}

In addition, pseudo label $p_i$ is used to extract visual concepts from the video's features. To obtain the representation $v_n$ of the $n^{th}$ visual concept, we pool the features of clips whose pseudo label is equal to $n$.

\begin{equation}
v_n=\frac{1}{\sum\limits_{i=1}^T\mathbf{1}(p_i=n)}\sum_{i=1}^T\mathbf{1}(p_i=n)\cdot x_i
\label{eq:visual_concept}
\end{equation}

The above equation holds if at least one clip is labeled as cluster $n$. If no clip corresponds to cluster $n$ in this video, we set $v_n=\vec{0}$.

Each visual concept represents a specific atomic action, and each fine-grained action can be modeled as a set of atomic actions. We connect the MIL classifier $\mathcal{S}$ to visual concepts $v=\{v_1,\cdots,v_N\}$ to learn the composition relationship between fine-grained actions and atomic actions. Given that the MIL classifier $\mathcal{S}$ maps clip features $x$ to their action prediction scores $s$, the last layer's weights $w$ in clip-level classifier $S$ naturally define class prototypes. Since visual concepts $v$ can also be viewed as an aggregated feature of clips, we propose to reinterpret the linear classifier weights $w$ on clip features $x$ as a nearest neighbor classifier on visual concepts $v$, to find relevant visual concepts for each fine-grained action. Specifically, we measure the relationship between fine classes and visual concepts through a distance function $\mathcal{D}$. $\mathcal{D}$ calculates the feature distance between visual concepts $v$ and the classification weights $w$ in MIL classifier $\mathcal{S}$.

\begin{equation}
d_n^{j}= \mathcal{D} (v_n, w^j)
\label{eq:distance}
\end{equation}

Intuitively, $d_n^{j}$ should be small if the $j^{th}$ fine-grained action consists of the atomic action represented by the $n^{th}$ visual concept $v_n$. We use cosine similarity in the distance function $\mathcal{D}$.

For each action $j$ in the video, we calculate $\mathcal{D}^j$, the distances between $w_j$ and all visual concepts, and then select the top $k$ visual concepts with the smallest distance ($d_n^{j}$), denoted as the set $TopK (\mathcal{D}^j)$. These $k$ visual concepts are then considered as atomic actions for fine action $j$. Thus we can generate fine-grained action $j$'s representation $e^j$ using these selected visual concepts' representation via a composition function $\mathcal{C}$.

\begin{equation}
  e^j = \mathcal{C} (~\{v_n | n \in TopK (\mathcal{D}^j) \}~)
  \label{eq:function_d}
\end{equation}

 Average pooling is used in the composition function $\mathcal{C}$. We then enforce the fine-grained action representation $e^j$ to be close to the class prototype $w^j$ by minimizing their distance measured by $\mathcal{D}$:

\begin{equation}
\mathcal{L}_{concept} = \mathcal{D} (e^j, w^j)
\label{eq:l_concept}
\end{equation}

$\mathcal{L}_{concept}$ is calculated only when the video contains fine action $j$.

\subsection{Coarse-to-Fine Semantic Hierarchy}
Due to the fine granularity in the label space, many actions are very close to each other with only minor differences.
Based on the commonality between actions, fine-grained actions can usually be grouped into a few coarse-grained hyper-categories, forming a semantic hierarchy. Actions within the same coarse category are visually more similar to each other, when compared to actions from different coarse categories (e.g., $cut~hair$ is visually closer to $brush~hair$ than $play~ tennis$). This semantic label hierarchy thus implies a useful prior of the feature distribution that helps visual concept learning. 

In \cref{subsec:visual_concept} we obtain the action representation $e^j $ composed by visual concepts through composition function $\mathcal{C}$. This formula models a hierarchical relationship between visual concepts and fine-grained actions. Similarly, a fine-to-coarse composition function $\mathcal{C'}$ can also be used to model the semantic relationship between fine-grained and coarse-grained actions. Specifically, for the $u^{th}$ coarse-grained category $O_u$ containing a set  of fine-grained classes, we compose the features of the fine-grained actions in $O_u$ to obtain the coarse-grained action representation $e_u'$:

\begin{equation}
  e_u' = \mathcal{C'} (~\{e^j | j \in O_u \}~)
  \label{eq:coarse}
\end{equation}

After obtaining coarse-grained action representation $e_u'$, we use the coarse-grained action label $y'$ to train a coarse-grained action classifier $\mathcal{S}'$ with binary cross entropy ($BCE$) loss with logits:

\begin{equation}
  \mathcal{L}_{coarse} = BCE(\mathcal{S}'(e_u'), y') 
  \label{eq:l_coarse}
\end{equation}

We combine these four losses to form the total loss $L$ used at training time, with $\lambda$ as weights.
\begin{equation}
  \mathcal{L}=\lambda_{1}\mathcal{L}_{mil}+\lambda_{2}\mathcal{L}_{pseudo}+\lambda_{3}\mathcal{L}_{concept}+\lambda_{4}\mathcal{L}_{coarse}
  \label{eq:loss}
\end{equation}

\subsection{Inference}

The inference process contains two steps. The first step is action classification. We run the MIL classifier $\mathcal{S}$ to obtain classification scores $s_i$ for each clip. Then for each action class $j$, we take the top $k$ clips which contain the highest activation scores for that class and average their scores. Then we threshold the average score to predict whether the video contains that action class or not.  

The second step is action detection. For each positive action class $j$, we calculate a threshold based on the class activation score sequence $s^j=\{s_1^j,s_2^j,\cdots,s_T^j\} $:

\begin{equation}
  {thresh}^j = Mean(s^j) + \alpha (Max(s^j) - Min(s^j) ) 
  \label{eq:detect}
\end{equation}

Clips with scores above the threshold ${thresh}^j$ are detected as actions, and we connect adjacent positive action clips to generate action segments.

\section{Experiments}

We test our model on two fine-grained video datasets, FineAction \cite{liu2021fineaction} and FineGym \cite{shao2020finegym}. In this section, we first introduce datasets and evaluation metrics. We also present implementation details. Then we discuss the main experiment results, followed by ablation studies on each of our components. In addition, we analyze the proposed visual concept learning with some qualitative results. 

\subsection{Datasets and Evaluation Metrics}

\textbf{FineAction}~\cite{liu2021fineaction} combines three existing datasets, YouTube8M~\cite{abu2016youtube}, Kinetics400~\cite{carreira2017quo}, FCVID~\cite{jiangfcvid}, and adds more videos crawled from the Internet. It has 8,440 videos with 57,752 action segments in the training set and 4,174 videos with 24,236 action segments in the validation set. The dataset contains a wide range of video contents including sports, household activities, personal care, etc. FineAction contains a three-level coarse-to-fine label hierarchy in its annotations. The three label levels contain 4, 14, 106 categories respectively and temporal annotations are available for every action label. The results reported by \cite{liu2021fineaction} is trained on the training set and tested on the validation set, so we follow the same setup. In addition to the 106-category fine-grained labels, we use the middle-level labels (14 classes) as the coarse-level labels in HAAN model.

For fair comparison in this dataset, we use the same evaluation metrics as in the FineAction dataset paper~\cite{liu2021fineaction}. Specifically, mean average precision (mAP) at temporal IoU thresholds from 0.5 to 0.95 with an interval of 0.05 is reported, as well as mAP@{0.5, 0.75, 0.95}.

\textbf{FineGym} \cite{shao2020finegym} is a dataset of gymnastics videos with three levels of annotations: $Events$, $Sets$ and $Elements$. An $event$ is a program routine of a player performing a whole set of actions. Within each $event$, actions are temporally annotated. Each action has two levels of labels, $sets$ (coarse) and $elements$ (fine). Two types of action detection can be done in FineGym: $event$ detection within a video, and $element$ detection within an $event$. The latter one is a fine-grained action detection task, thus we use it to conduct our experiments. Our experiments use FineGym-99, which has 14 coarse labels and 99 fine labels.

We find that the data split introduced in the original FineGym paper \cite{shao2020finegym} is not suitable for action detection experiments, because under that data split, actions in the training set and the validation set can come from the same video. As a result, 18\% of the validation videos are actually seen by the model during training. To avoid the data leakage which can lead to unrealistically high results during testing, we propose a new train-val data split in FineGym using an iterative sampling approach with corresponding ratio control.

The training-to-validation sample ratio in the original split of FineGym is 75 to 25. In the iterative sampling approach, our goal is to also select 75\% of the total samples as training samples, and at the same time maintain the 75-to-25 ratio in each action class as much as possible. Since each video in FineGym might contain more than one action class, we follow the greedy search idea to generate the new data split. We first add a minimum number of videos to both training and validation sets so that both of them contain all action classes. 
Then we find the action class with the least per-class sample ratio in the training set, and sample a video containing that class into the training set. 
We repeat this process until each action class in the training set contains at least 75\% samples in that class. We run the sampling algorithm for 100 times, and pick the resulting data split whose sample distribution is the closest to our criterion with the approximate 75-to-25 ratios in both overall and per-class sample distribution. The new data split contains 3,775 videos with 26,866 action segments in the training set, and 1,193 videos with 7,975 action segments in the validation set. The training-to-validation sample ratios of each action class range from 75\% to 81\%.

Average mAP at temporal IoU thresholds from 0.1 to 0.5 with an interval of 0.05 is reported in this data split, as well as mAP@{0.1, 0.2, 0.3, 0.4, 0.5}.

\begin{table}[t]
  \setlength{\tabcolsep}{6pt}
  \centering
  \caption{Results on FineAction dataset. The avg.mAP refers to the average of mAPs at temporal IoU thresholds ranging from 0.5 to 0.95 with an interval of 0.05. The fully-supervised BMN model~\cite{lin2019bmn}'s results are from \cite{liu2021fineaction}. We run previous weakly-supervised models \cite{paul2018w,pardo2021refineloc,lee2021weakly,ma2021weakly,narayan2021d2} using publicly available code. Ablation studies of our HAAN model with different losses enabled are also presented}
  \begin{tabular}{@{}l|l|ccc|c@{}}
    \hline
    \multicolumn{2}{@{}l|}{\multirow{2}{*}{Methods}} & \multicolumn{3}{c|}{mAP@ $\tau$ } & \multirow{2}{*}{avg.mAP} \\
    \multicolumn{2}{@{}l|}{} & 0.5 & 0.75 & 0.95 &\\
    \hline
    \multirow{5}{*}{Prior} & BMN (fully-supervised) (2019)
    \cite{lin2019bmn} & 14.44 & 8.92 & 3.12 & 9.25\\
    \cline{2-6}
     & W-TALC (2018) \cite{paul2018w}  & 6.18 & 3.15& 0.83 & 3.45\\
     & RefineLoc (2021) \cite{pardo2021refineloc} & 5.93 &2.55 & 0.96& 3.02\\
     & WTAL-UM (2021) \cite{lee2021weakly}  & 6.65 &3.23 &0.95 &3.64 \\
     & ASL (2021) \cite{ma2021weakly}  & 6.79 & 2.68 & 0.81 & 3.30\\
     & D2-Net (2021) \cite{narayan2021d2} & 6.75 & 3.02 & 0.82 & 3.35\\
    \hline
    \multirow{4}{*}{Ours} & $\mathcal{L}_{mil}$ & 5.74 & 2.29 & 0.39 & 2.72 \\
    & $\mathcal{L}_{mil}$+$\mathcal{L}_{pseudo}$ & 6.40 & 2.73 & 0.61 & 3.18 \\
    & $\mathcal{L}_{mil}$+$\mathcal{L}_{pseudo}$+$\mathcal{L}_{concept}$ & 6.57 & 3.27 & 0.83 & 3.57 \\
    & $\mathcal{L}_{mil}$+$\mathcal{L}_{pseudo}$+$\mathcal{L}_{concept}$+$\mathcal{L}_{coarse}$ & \textbf{7.05} & \textbf{3.95} & \textbf{1.14} & \textbf{4.10} \\
    \hline
  \end{tabular}
  
  \label{tab:results:fineaction}
\end{table}

\begin{table}[t]
\setlength{\tabcolsep}{3pt}
  \centering
  \caption{Results on FineGym dataset. The fully-supervised SSN model's results are reported on the new data split. We also run previous weakly-supervised models~\cite{paul2018w,pardo2021refineloc,lee2021weakly,ma2021weakly,narayan2021d2} using publicly available code. Ablation results of our HAAN model with different losses enabled are also presented}
  \begin{tabular}{@{}l|l|ccccc|c@{}}
    \hline
    \multicolumn{2}{@{}l|}{\multirow{2}{*}{Methods}} & \multicolumn{5}{c|}{mAP@ $\tau$ } & \multirow{2}{*}{avg.mAP} \\
    \multicolumn{2}{@{}l|}{} & 0.1 & 0.2 & 0.3 & 0.4 & 0.5 &\\
    \hline
    \multirow{5}{*}{Prior} & SSN (fully-supervised) (2017) \cite{zhao2017temporal} & 20.23 & 19.29&17.12 &14.38  & 11.45&16.61\\
    \cline{2-8}
    & W-TALC (2018) \cite{paul2018w} & 8.85 & 7.32 & 6.24 &4.95 &3.15 &6.03\\
    & RefineLoc (2021) \cite{pardo2021refineloc} & 6.67 &4.63 &4.15 &3.86 & 3.72& 4.54\\
    & WTAL-UM (2021) \cite{lee2021weakly} & 9.45 &8.63 &5.10 &4.34 &3.05 & 6.11\\
    & ASL (2021) \cite{ma2021weakly} & 9.33 & 7.92 & 5.45& 3.67&2.24 & 5.74\\
    & D2-Net (2021) \cite{narayan2021d2} & 9.46 & 8.67 & 5.21 & 4.22 & 2.65 & 6.04\\
    \hline
    \multirow{4}{*}{Ours} & $\mathcal{L}_{mil}$ &7.78 & 7.23& 5.92& 4.12& 2.98 & 5.61\\
    & $\mathcal{L}_{mil}$+$\mathcal{L}_{pseudo}$ & 8.56& 7.50& 5.83& 4.60& 3.45 & 6.09 \\
    & $\mathcal{L}_{mil}$+$\mathcal{L}_{pseudo}$+$\mathcal{L}_{concept}$ & 9.36  & 8.28 & 6.99 & 5.25 & 3.91 & 6.95 \\
    & $\mathcal{L}_{mil}$+$\mathcal{L}_{pseudo}$+$\mathcal{L}_{concept}$+$\mathcal{L}_{coarse}$ & \textbf{10.79} & \textbf{9.62} & \textbf{7.65} & \textbf{6.16} & \textbf{4.16} & \textbf{7.67} \\
    \hline
  \end{tabular}
  
  \label{tab:results:finegym}
\end{table}

\begin{table}[t]
  \setlength{\tabcolsep}{6pt}
  \centering
  \caption{mAP@0.5 results for different clustering methods on FineAction dataset using the HAAN model version with $\mathcal{L}_{mil}$+$\mathcal{L}_{pseudo}$ supervision}
  \begin{tabular}{c|cccc}
    \hline
     \multirow{2}{*}{Methods}& \multicolumn{4}{c}{Number of Clusters} \\
     & 250 & 500 & 750 & 1000 \\
    \hline
     GMM~\cite{reynolds2009gaussian} & 6.15 & 6.14 & 6.20 & 6.36\\
     Birch~\cite{zhang1996birch} & 6.25 & 6.31 & 6.19 & 6.31\\
     K-means \cite{macqueen1967classification} & 6.31 & 6.40 & 6.25 & 6.26\\
    \hline
  \end{tabular}
  
  \label{tab:results:clustering}
\end{table}

\begin{table}[t]
  \setlength{\tabcolsep}{6pt}
  \centering
  \caption{Results of different distance calculations on FineAction dataset using the HAAN model version with $\mathcal{L}_{mil}$+$\mathcal{L}_{pseudo}$+$\mathcal{L}_{concept}$ supervision}
  \begin{tabular}{c|ccc|c}
    \hline
   Method & mAP@0.5 & mAP@0.75 & mAP@0.95 & avg.mAP\\
    \hline
     Euclidean & 5.92 & 3.03 & 0.81 & 3.26\\
     cosine & 6.57 & 3.27 & 0.83 & 3.57\\
    \hline
  \end{tabular}
  
  \label{tab:results:mse}
\end{table}

\subsection{Implementation Details}
Our model is implemented in PyTorch. We use two fully connected layers on top of the two-stream Inception3D (I3D) backbone~\cite{carreira2017quo} with RGB and Optical Flow inputs to form the feature encoder $\mathcal{E}$, following the standard practice in weakly-supervised temporal action detection, e.g.~\cite{nguyen2018weakly}. We use one fully connected layer for MIL classifier $\mathcal{S}$, and two fully connected layers for self-supervised visual concept classifier $\mathcal{F}$. The number of visual concepts $N$ is set as $500$, and each class retains top $k=5$ relevant visual concepts. Fine-to-coarse composition function $\mathcal{C'}$ is max pooling for FineAction and mean pooling for FineGym. We use Adam optimizer with learning rate $\beta=3\times10^{-5}$. The weights for each loss are: $\lambda_1=1,\lambda_2=0.001,\lambda_3=0.01,\lambda_4=1$. The $\alpha$ in \cref{eq:detect} is $-0.8$ for FineAction and $0.1$ for FineGym.

\subsection{Main Results}
We compare our HAAN model to previous weakly-supervised temporal action detection models~\cite{paul2018w,pardo2021refineloc,ma2021weakly,lee2021weakly,narayan2021d2}, and results on the two datasets are shown in~\cref{tab:results:fineaction,tab:results:finegym}. Results of fully-supervised models (BMN or SSN) are also listed as reference. In FineGym dataset, we re-run the fully-supervised algorithm SSN \cite{zhao2017temporal} on the new data split, and the new results shown in~\cref{tab:results:finegym} are worse than the results reported in FineGym paper~\cite{shao2020finegym} as expected, because the original data split has train-test video data leakage problem.
Overall, fully-supervised models also struggle on both datasets with low performance, demonstrating the difficulty of temporal action detection on fine-grained datasets.

Our HAAN model outperforms the best weakly-supervised detection model WTAL-UM \cite{lee2021weakly} by 12\% on FineAction and 25\% on FineGym in average mAP relatively. We also find that more advanced weakly models ASL \cite{ma2021weakly} and RefineLoc \cite{pardo2021refineloc} are not superior to the older weakly model W-TALC \cite{paul2018w} in fine-grained videos. This again indicates the big difference between fine-grained action datasets and general action datasets. Models that perform well on THUMOS-14 and ActivityNet-1.2 may not perform well on fine-grained video datasets.

Results in \cref{tab:results:fineaction} are reported on the validation set of FineAction. We also submit our HAAN model's predictions on the withheld test set to the FineAction competition leaderboard\footnote{\url{https://competitions.codalab.org/competitions/32363}}, and get the average mAP of 4.48 on the test set with hidden labels, which is close to the average mAP of 4.10 on the validation set.

For ablation study, we also include four different versions of our model with different losses enabled in \cref{tab:results:fineaction,tab:results:finegym}. It shows that each component in our HAAN model has a considerable contribution to the detection performance.

\subsection{Visual Concept Learning Analysis}
\label{sec:vc-learning}
We use K-means clustering to obtain visual concepts, and then build connections with fine-grained action classes by minimizing the cosine distance between visual concepts and the MIL classifier's weights. Here we show the ablation results of different clustering methods and distance calculations in visual concept learning.

For the clustering method to generate pseudo labels, we explore Gaussian Mixture Model (GMM)~\cite{reynolds2009gaussian} and Birch~\cite{zhang1996birch} in addition to K-means, and test on different numbers of clusters. 
The HAAN model version with $\mathcal{L}_{mil}$+$\mathcal{L}_{pseudo}$ supervision is used in this ablation to better reflect the effect of different clustering methods. The ablated mAP@0.5 results on FineAction dataset are shown in~\cref{tab:results:clustering}. Among all the clustering methods, K-means with 500 clusters achieves the best result.

To build connection between visual concepts and fine-grained action classes, distance metric is used, e.g. in the mapping function $\mathcal{D}$ (\cref{eq:distance}) between visual concepts and fine-grained actions, and the auxiliary loss function for fine-grained action representation learning based on visual concepts (\cref{eq:l_concept}). We experiment with Euclidean distance and cosine similarity in the HAAN model version with $\mathcal{L}_{mil}$+$\mathcal{L}_{pseudo}$+$\mathcal{L}_{concept}$ supervision, and ablation results are shown in \cref{tab:results:mse}. Cosine similarity outperforms Euclidean distance generally in our model.

\subsection{Qualitative Results}

\begin{figure}[t]
  \centering
   \includegraphics[width=0.4\linewidth]{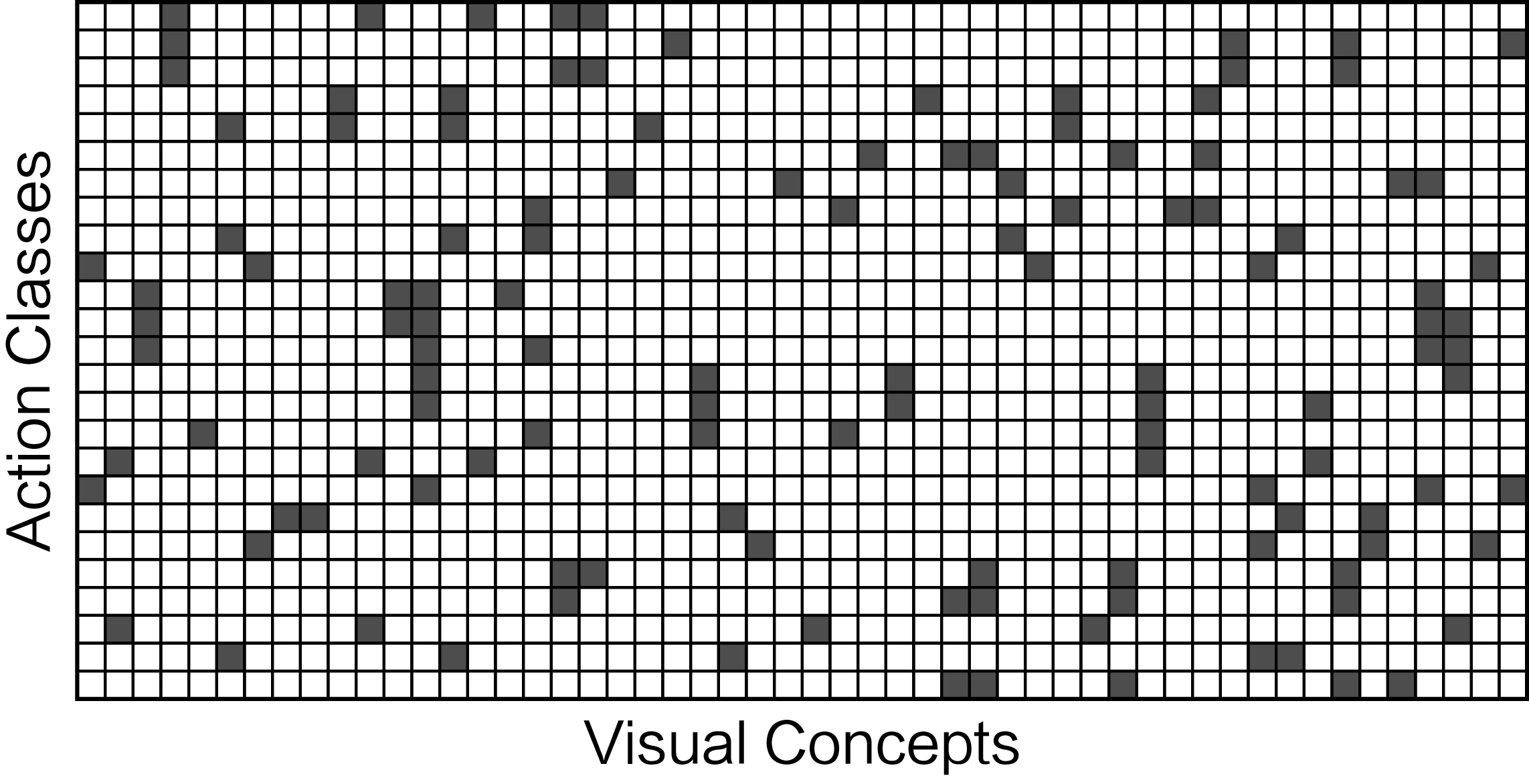}
   \caption{Visualization of relevant visual concepts for each fine-grained action class in the \textit{Uneven Bars} event from FineGym dataset}
   \label{fig:concepts-actions}
\end{figure}

\begin{figure*}[t]
  \centering
   \includegraphics[width=0.95\linewidth]{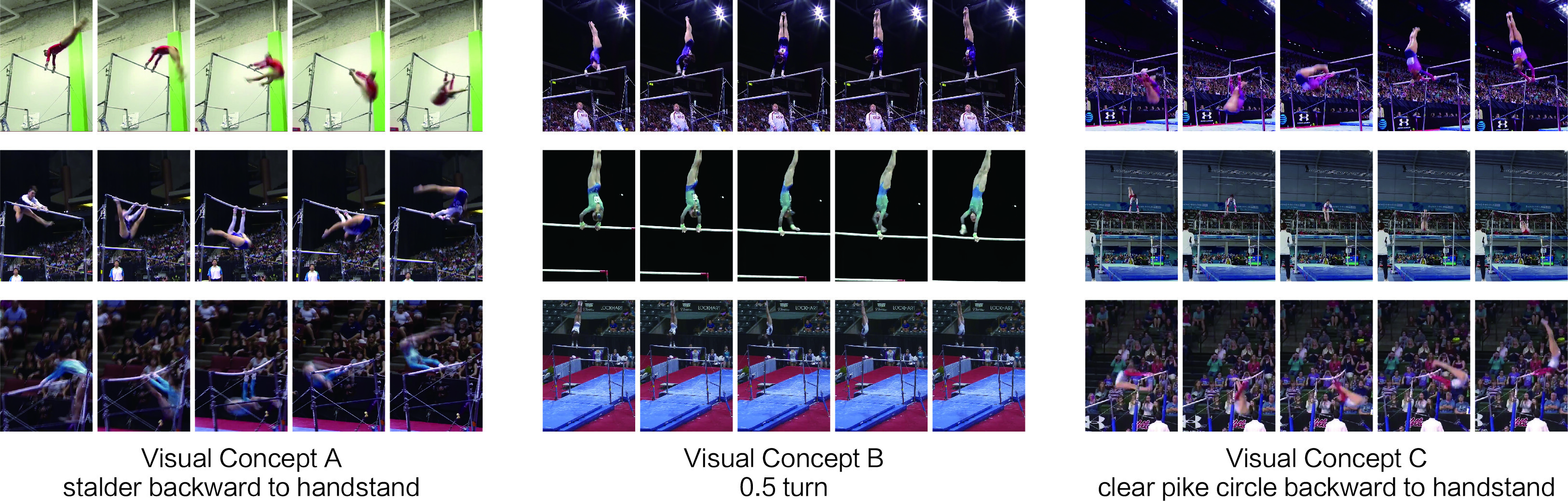}
   \caption{Three example visual concepts learned in our model representing atomic actions}
   \label{fig:concepts-example}
\end{figure*}

\begin{figure*}[t]
  \centering
   \includegraphics[width=0.95\linewidth]{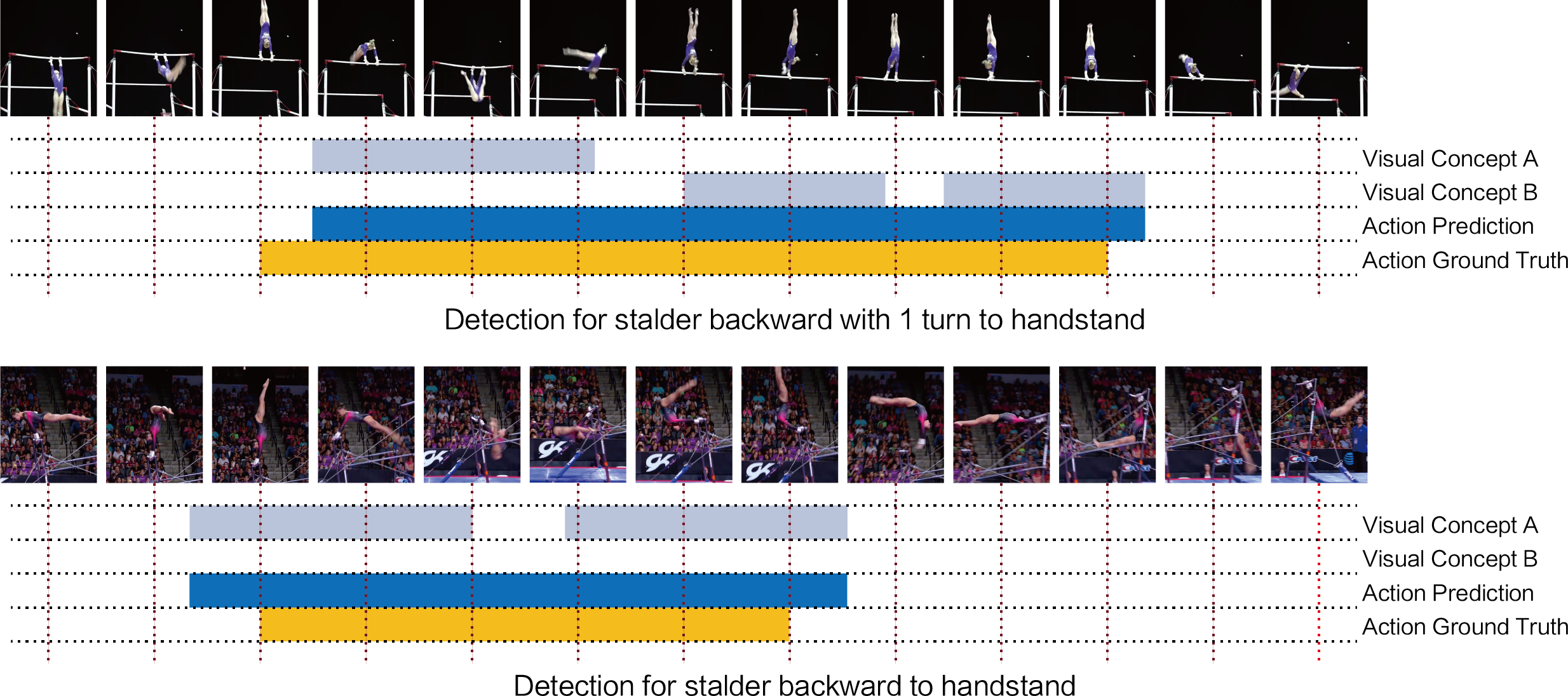}
   \caption{Two example action detections from our HAAN model. Visual concepts A and B correspond to those visualized in \cref{fig:concepts-example}}
   \label{fig:detection}
\end{figure*}

Our model assumes that fine-grained actions share common visual concepts (a.k.a. atomic actions) and vary in certain visual concepts. \cref{fig:concepts-actions} shows the visual concepts that each fine-grained action class is most related to in the \textit{Uneven Bars} event from FineGym dataset. The appearance pattern of visual concepts indeed demonstrates that there exists visual concepts that are shared by actions, and every action can be represented by a unique combination of visual concepts.

To better understand what the visual concepts are, we plot some examples of visual concepts in \cref{fig:concepts-example}, and the specific atomic action label below each visual concept (Visual concept A, B, C) is generated from our observation for illustration purpose. Each visual concept represents an atomic action. Visual concept A represents atomic action \textit{stalder backward to handstand} where the athlete starts from handstand phase, circles around the bar with legs wide apart, and moves backward to handstand. Visual concept B represents atomic action \textit{0.5 turn} on handstand. Visual concept C represents \textit{clear pike circle backward to handstand} with two legs together. Visual concepts A and C differ in whether the athlete's legs are wide apart or together in the circle.
Also, visual concept B can occur following visual concept A to form action \textit{stalder backward with 0.5/1 turn to handstand}, or following visual concept C to form action \textit{clear pike circle backward with 0.5/1 turn to handstand}.

Our model also learns to compose visual concepts into fine-grained actions. As shown in \cref{fig:detection}, our model successfully detects the sequence of visual concept A followed by two visual concepts B as fine-grained action~\textit{stalder backward with 1 turn to handstand}, and the sequence of only visual concepts A as fine-grained action~\textit{stalder backward to handstand}.

\section{Conclusion}
We propose Hierarchical Atomic Action Network (HAAN) to address weakly-supervised temporal action detection in fine-grained videos for the first time. HAAN automatically discovers the visual concepts to capture the fine-grained action details, utilizing clustering-based self-supervised learning and the coarse-to-fine action label hierarchy. Experiment results demonstrate that HAAN outperforms state-of-the-art weakly-supervised methods on two large-scale fine-grained video datasets, FineAction and FineGym.

\clearpage

\bibliographystyle{splncs04}
\bibliography{egbib}

\begin{thebibliography}{10}
\providecommand{\url}[1]{\texttt{#1}}
\providecommand{\urlprefix}{URL }
\providecommand{\doi}[1]{https://doi.org/#1}

\bibitem{abu2016youtube}
Abu-El-Haija, S., Kothari, N., Lee, J., Natsev, P., Toderici, G., Varadarajan,
  B., Vijayanarasimhan, S.: Youtube-8m: A large-scale video classification
  benchmark. arXiv preprint arXiv:1609.08675  (2016)

\bibitem{caba2015activitynet}
Caba~Heilbron, F., Escorcia, V., Ghanem, B., Carlos~Niebles, J.: Activitynet: A
  large-scale video benchmark for human activity understanding. In: Proceedings
  of the IEEE conference on computer vision and pattern recognition. pp.
  961--970 (2015)

\bibitem{carbonneau2018multiple}
Carbonneau, M.A., Cheplygina, V., Granger, E., Gagnon, G.: Multiple instance
  learning: A survey of problem characteristics and applications. Pattern
  Recognition  \textbf{77},  329--353 (2018)

\bibitem{carreira2017quo}
Carreira, J., Zisserman, A.: Quo vadis, action recognition? a new model and the
  kinetics dataset. In: proceedings of the IEEE Conference on Computer Vision
  and Pattern Recognition. pp. 6299--6308 (2017)

\bibitem{chen2019looks}
Chen, C., Li, O., Tao, D., Barnett, A., Rudin, C., Su, J.K.: This looks like
  that: deep learning for interpretable image recognition. Advances in neural
  information processing systems  \textbf{32} (2019)

\bibitem{chen2016infogan}
Chen, X., Duan, Y., Houthooft, R., Schulman, J., Sutskever, I., Abbeel, P.:
  Infogan: Interpretable representation learning by information maximizing
  generative adversarial nets. In: Proceedings of the 30th International
  Conference on Neural Information Processing Systems. pp. 2180--2188 (2016)

\bibitem{damen2022rescaling}
Damen, D., Doughty, H., Farinella, G.M., Furnari, A., Kazakos, E., Ma, J.,
  Moltisanti, D., Munro, J., Perrett, T., Price, W., et~al.: Rescaling
  egocentric vision: collection, pipeline and challenges for epic-kitchens-100.
  International Journal of Computer Vision  \textbf{130}(1),  33--55 (2022)

\bibitem{DIETTERICH199731}
Dietterich, T.G., Lathrop, R.H., Lozano-Pérez, T.: Solving the multiple
  instance problem with axis-parallel rectangles. Artificial Intelligence
  \textbf{89}(1),  31--71 (1997)

\bibitem{fathi2011learning}
Fathi, A., Ren, X., Rehg, J.M.: Learning to recognize objects in egocentric
  activities. In: CVPR 2011. pp. 3281--3288. IEEE (2011)

\bibitem{gaidon2013temporal}
Gaidon, A., Harchaoui, Z., Schmid, C.: Temporal localization of actions with
  actoms. IEEE transactions on pattern analysis and machine intelligence
  \textbf{35}(11),  2782--2795 (2013)

\bibitem{ghoddoosian2022hierarchical}
Ghoddoosian, R., Sayed, S., Athitsos, V.: Hierarchical modeling for task
  recognition and action segmentation in weakly-labeled instructional videos.
  In: Proceedings of the IEEE/CVF Winter Conference on Applications of Computer
  Vision. pp. 1922--1932 (2022)

\bibitem{ghorbani2019automating}
Ghorbani, A., Wexler, J., Kim, B.: Automating interpretability: Discovering and
  testing visual concepts learned by neural networks. ArXiv
  \textbf{abs/1902.03129} (2019)

\bibitem{higgins2017scan}
Higgins, I., Sonnerat, N., Matthey, L., Pal, A., Burgess, C.P., Bosnjak, M.,
  Shanahan, M., Botvinick, M., Hassabis, D., Lerchner, A.: Scan: Learning
  hierarchical compositional visual concepts. arXiv preprint arXiv:1707.03389
  (2017)

\bibitem{MIL_summary}
Ilse, M., Tomczak, J., Welling, M.: Attention-based deep multiple instance
  learning. In: Dy, J., Krause, A. (eds.) Proceedings of the 35th International
  Conference on Machine Learning. Proceedings of Machine Learning Research,
  vol.~80, pp. 2127--2136. PMLR (10--15 Jul 2018)

\bibitem{ji2020action}
Ji, J., Krishna, R., Fei-Fei, L., Niebles, J.C.: Action genome: Actions as
  compositions of spatio-temporal scene graphs. In: Proceedings of the IEEE/CVF
  Conference on Computer Vision and Pattern Recognition. pp. 10236--10247
  (2020)

\bibitem{jiang2014thumos}
Jiang, Y.G., Liu, J., Zamir, A.R., Toderici, G., Laptev, I., Shah, M.,
  Sukthankar, R.: Thumos challenge: Action recognition with a large number of
  classes (2014)

\bibitem{jiangfcvid}
Jiang, Y.G., Wu, Z., Wang, J., Xue, X., Chang, S.F.: Exploiting feature and
  class relationships in video categorization with regularized deep neural
  networks. {IEEE} Transactions on Pattern Analysis and Machine Intelligence
  \textbf{40}(2),  352--364 (2018)

\bibitem{kim2018interpretability}
Kim, B., Wattenberg, M., Gilmer, J., Cai, C., Wexler, J., Viegas, F., et~al.:
  Interpretability beyond feature attribution: Quantitative testing with
  concept activation vectors (tcav). In: International conference on machine
  learning. pp. 2668--2677. PMLR (2018)

\bibitem{lee2021weakly}
Lee, P., Wang, J., Lu, Y., Byun, H.: Weakly-supervised temporal action
  localization by uncertainty modeling. In: AAAI Conference on Artificial
  Intelligence. vol.~2 (2021)

\bibitem{lillo2014discriminative}
Lillo, I., Soto, A., Carlos~Niebles, J.: Discriminative hierarchical modeling
  of spatio-temporally composable human activities. In: Proceedings of the IEEE
  conference on computer vision and pattern recognition. pp. 812--819 (2014)

\bibitem{lin2019bmn}
Lin, T., Liu, X., Li, X., Ding, E., Wen, S.: Bmn: Boundary-matching network for
  temporal action proposal generation. In: Proceedings of the IEEE/CVF
  International Conference on Computer Vision. pp. 3889--3898 (2019)

\bibitem{liu2019completeness}
Liu, D., Jiang, T., Wang, Y.: Completeness modeling and context separation for
  weakly supervised temporal action localization. In: Proceedings of the
  IEEE/CVF Conference on Computer Vision and Pattern Recognition. pp.
  1298--1307 (2019)

\bibitem{liu2021fineaction}
Liu, Y., Wang, L., Ma, X., Wang, Y., Qiao, Y.: Fineaction: A fine-grained video
  dataset for temporal action localization. arXiv preprint arXiv:2105.11107
  (2021)

\bibitem{luo2020weakly}
Luo, Z., Guillory, D., Shi, B., Ke, W., Wan, F., Darrell, T., Xu, H.:
  Weakly-supervised action localization with expectation-maximization
  multi-instance learning. In: European conference on computer vision. pp.
  729--745. Springer (2020)

\bibitem{ma2021weakly}
Ma, J., Gorti, S.K., Volkovs, M., Yu, G.: Weakly supervised action selection
  learning in video. In: Proceedings of the IEEE/CVF Conference on Computer
  Vision and Pattern Recognition. pp. 7587--7596 (2021)

\bibitem{deform}
Mac, K.N.C., Joshi, D., Yeh, R.A., Xiong, J., Feris, R.S., Do, M.N.: Learning
  motion in feature space: Locally-consistent deformable convolution networks
  for fine-grained action detection. In: Proceedings of the IEEE/CVF
  International Conference on Computer Vision. pp. 6282--6291 (2019)

\bibitem{macqueen1967classification}
MacQueen, J.: Classification and analysis of multivariate observations. In: 5th
  Berkeley Symp. Math. Statist. Probability. pp. 281--297 (1967)

\bibitem{random_field}
Mavroudi, E., Bhaskara, D., Sefati, S., Ali, H., Vidal, R.: End-to-end
  fine-grained action segmentation and recognition using conditional random
  field models and discriminative sparse coding. In: 2018 IEEE Winter
  Conference on Applications of Computer Vision (WACV). pp. 1558--1567. IEEE
  (2018)

\bibitem{narayan2021d2}
Narayan, S., Cholakkal, H., Hayat, M., Khan, F.S., Yang, M.H., Shao, L.:
  D2-net: Weakly-supervised action localization via discriminative embeddings
  and denoised activations. In: Proceedings of the IEEE/CVF International
  Conference on Computer Vision. pp. 13608--13617 (2021)

\bibitem{narayan20193c}
Narayan, S., Cholakkal, H., Khan, F.S., Shao, L.: 3c-net: Category count and
  center loss for weakly-supervised action localization. In: Proceedings of the
  IEEE/CVF International Conference on Computer Vision. pp. 8679--8687 (2019)

\bibitem{nguyen2018weakly}
Nguyen, P., Liu, T., Prasad, G., Han, B.: Weakly supervised action localization
  by sparse temporal pooling network. In: Proceedings of the IEEE Conference on
  Computer Vision and Pattern Recognition. pp. 6752--6761 (2018)

\bibitem{ni2014multiple}
Ni, B., Paramathayalan, V.R., Moulin, P.: Multiple granularity analysis for
  fine-grained action detection. In: Proceedings of the IEEE Conference on
  Computer Vision and Pattern Recognition. pp. 756--763 (2014)

\bibitem{pardo2021refineloc}
Pardo, A., Alwassel, H., Caba, F., Thabet, A., Ghanem, B.: Refineloc: Iterative
  refinement for weakly-supervised action localization. In: Proceedings of the
  IEEE/CVF Winter Conference on Applications of Computer Vision. pp. 3319--3328
  (2021)

\bibitem{paul2018w}
Paul, S., Roy, S., Roy-Chowdhury, A.K.: W-talc: Weakly-supervised temporal
  activity localization and classification. In: Proceedings of the European
  Conference on Computer Vision (ECCV). pp. 563--579 (2018)

\bibitem{FineBaseball}
Piergiovanni, A.J., Ryoo, M.S.: Fine-grained activity recognition in baseball
  videos. 2018 IEEE/CVF Conference on Computer Vision and Pattern Recognition
  Workshops (CVPRW) pp. 1821--18218 (2018)

\bibitem{reynolds2009gaussian}
Reynolds, D.A.: Gaussian mixture models. Encyclopedia of biometrics
  \textbf{741},  659--663 (2009)

\bibitem{richard2017weakly}
Richard, A., Kuehne, H., Gall, J.: Weakly supervised action learning with rnn
  based fine-to-coarse modeling. In: Proceedings of the IEEE conference on
  Computer Vision and Pattern Recognition. pp. 754--763 (2017)

\bibitem{rohrbach2012database}
Rohrbach, M., Amin, S., Andriluka, M., Schiele, B.: A database for fine grained
  activity detection of cooking activities. In: 2012 IEEE Conference on
  Computer Vision and Pattern Recognition. pp. 1194--1201. IEEE (2012)

\bibitem{shao2020finegym}
Shao, D., Zhao, Y., Dai, B., Lin, D.: Finegym: A hierarchical video dataset for
  fine-grained action understanding. In: Proceedings of the IEEE/CVF Conference
  on Computer Vision and Pattern Recognition. pp. 2616--2625 (2020)

\bibitem{shou2018autoloc}
Shou, Z., Gao, H., Zhang, L., Miyazawa, K., Chang, S.F.: Autoloc:
  Weakly-supervised temporal action localization in untrimmed videos. In:
  Proceedings of the European Conference on Computer Vision (ECCV). pp.
  154--171 (2018)

\bibitem{singh2016multi}
Singh, B., Marks, T.K., Jones, M., Tuzel, O., Shao, M.: A multi-stream
  bi-directional recurrent neural network for fine-grained action detection.
  In: Proceedings of the IEEE conference on computer vision and pattern
  recognition. pp. 1961--1970 (2016)

\bibitem{stein2013combining}
Stein, S., McKenna, S.J.: Combining embedded accelerometers with computer
  vision for recognizing food preparation activities. In: Proceedings of the
  2013 ACM international joint conference on Pervasive and ubiquitous
  computing. pp. 729--738 (2013)

\bibitem{sun2015temporal}
Sun, C., Shetty, S., Sukthankar, R., Nevatia, R.: Temporal localization of
  fine-grained actions in videos by domain transfer from web images. In:
  Proceedings of the 23rd ACM international conference on Multimedia. pp.
  371--380 (2015)

\bibitem{wang2017untrimmednets}
Wang, L., Xiong, Y., Lin, D., Van~Gool, L.: Untrimmednets for weakly supervised
  action recognition and detection. In: Proceedings of the IEEE conference on
  Computer Vision and Pattern Recognition. pp. 4325--4334 (2017)

\bibitem{continuation_learning}
Whitney, W.F., Chang, M., Kulkarni, T., Tenenbaum, J.B.: Understanding visual
  concepts with continuation learning. arXiv preprint arXiv:1602.06822  (2016)

\bibitem{yuan2019marginalized}
Yuan, Y., Lyu, Y., Shen, X., Tsang, I., Yeung, D.Y.: Marginalized average
  attentional network for weakly-supervised learning. In: ICLR 2019-Seventh
  International Conference on Learning Representations (2019)

\bibitem{zhang1996birch}
Zhang, T., Ramakrishnan, R., Livny, M.: Birch: an efficient data clustering
  method for very large databases. ACM sigmod record  \textbf{25}(2),  103--114
  (1996)

\bibitem{zhao2017temporal}
Zhao, Y., Xiong, Y., Wang, L., Wu, Z., Tang, X., Lin, D.: Temporal action
  detection with structured segment networks. In: Proceedings of the IEEE
  International Conference on Computer Vision. pp. 2914--2923 (2017)

\end{thebibliography}

\end{document}